%% file: arxiv.tex
\documentclass[11pt]{article}

\usepackage[margin=0.68in]{geometry}
\usepackage{times}
\usepackage[authoryear,round]{natbib}
\input{math_commands.tex}

\usepackage{amsmath,amssymb,booktabs,multirow,graphicx,float}
\usepackage{hyperref}
\usepackage{url}
\usepackage{xcolor}
\usepackage{microtype}
\usepackage{xspace}

\title{NGN: Learning Neural Network Size as a \mbox{Differentiable} Count}
\author{%
  Li Lixing \\
  Cornell University \\
  Ithaca, NY 14853 \\
  \texttt{ll963@cornell.edu} \\
}
\date{}
\hypersetup{
  pdftitle={NGN: Learning Neural Network Size as a Differentiable Count},
  pdfauthor={Li Lixing}
}

\newcommand{\method}{\textsc{NGN}\xspace}

\newcommand{\paperbibliographystyle}{plainnat}
\newcommand{\beginpaperfigure}{\begin{figure}[!ht]}
\newcommand{\beginpapertable}{\begin{table}[!ht]}
\renewcommand{\softplus}{\operatorname{softplus}}
\renewcommand{\sigmoid}{\operatorname{sigmoid}}

\begin{document}
\input{paper_content.tex}
\end{document}

%% file: math_commands.tex
\usepackage{amsmath,amsfonts,bm}

\def\eqref#1{equation~\ref{#1}}
\def\1{\bm{1}}

\DeclareMathAlphabet{\mathsfit}{\encodingdefault}{\sfdefault}{m}{sl}
\SetMathAlphabet{\mathsfit}{bold}{\encodingdefault}{\sfdefault}{bx}{n}

\newcommand{\sigmoid}{\sigma}
\newcommand{\softplus}{\zeta}

%% file: paper_content.tex
\maketitle

\begin{abstract}
  Neural network size is usually chosen before training, separating architecture
  selection from weight optimization. We introduce the Neurogenesis Network
  (NGN), a differentiable parameterization for learning how many ordered
  structural components a model should use. For each
  ordered component group, one learnable boundary selects an active prefix while the
  model parameters are trained. The boundary can grow from a compact
  initialization and can be deployed by discarding components beyond the learned
  boundary. Controlled experiments examine convergence of the learned boundary,
  the performance of deployed prefixes, and comparisons with fixed-size models
  and alternative approaches to learning capacity. We then apply the same
  mechanism to MLPs, convolutional and graph networks, Transformers, state-space
  models, LoRA, and adapters. Across these settings, deploying only the learned
  prefix usually changes performance little, and the selected architectures
  perform similarly to fixed models trained at the same size. These results show
  that structural capacity can be optimized directly as a count.
\end{abstract}

\section{Introduction}

The size of a neural network is usually fixed before its weights are learned.
We choose its widths, depths, and other structural dimensions, then
train the resulting model. Selecting these quantities usually requires expertise or
repeated training across candidate sizes. Because useful capacity depends on the
data and objective, model size could instead be optimized alongside the weights.

Prior work on learning or selecting model size falls into four groups.
Post-training pruning removes components from an already trained model
\citep{frantar2023sparsegpt,ashkboos2024slicegpt}. Learned sparsification instead
selects components within a fixed, overcomplete model
\citep{louizos2018l0,ding2023sora}. Flexible computation models either support
multiple deployment sizes or vary computation amount by input
\citep{valipour2023dylora,graves2016act}. Finally, our approach belongs most
closely to joint structure learning, in which active capacity changes during
training \citep{nazaret2022udn,errica2026awnn}.

We propose the Neurogenesis Network (\method), a way to learn \emph{how many}
ordered components a model needs together with the components themselves. For candidate functions
$f_0,f_1,\ldots$, the deployed model has the form
\begin{equation}
  y(x)=\sum_{k=0}^{\infty} w_k g_k(t) f_k(x),
  \qquad g_k(t)\in\{0,1\},
  \label{eq:intro}
\end{equation}
where the fixed coefficient $w_k$ sets the scale of component $k$ and $g_k(t)$
is controlled by a shared boundary $t$, so the active components form a prefix.
\method learns one boundary for each ordered component group. The same representation can therefore control different
forms of model capacity. The active prefix can grow from a small initialization
during training; afterward, components beyond the learned boundary can simply be
discarded.

Across controlled regression and experiments in vision, graph learning, language
modeling, and parameter-efficient fine-tuning, \method learns task-dependent
sizes for a range of structural components. Discarding components beyond the
learned boundary usually changes performance little, and the resulting
architectures perform similarly to fixed models trained at the same size.

Our contributions are:
\begin{itemize}
  \item We introduce \method, a parameterization that represents model size as a
        differentiable count over an ordered sequence of components and learns
        that count jointly with the model weights.
  \item We conduct controlled analyses of how the learned count behaves, when it
        is stable and deployable, and how it compares with fixed-size models and
        alternative approaches to learning capacity.
  \item We demonstrate the same mechanism across diverse model families, tasks,
        and structural axes, including heterogeneous allocations across layers.
\end{itemize}

\section{Related Work}
\label{sec:related}

\paragraph{Post-training pruning.}
These methods begin from a pretrained model, estimate importance, and
remove weights or structures under a specified budget. SparseGPT uses approximate
second-order information for layer-wise reconstruction
\citep{frantar2023sparsegpt}, while Wanda ranks weights using their magnitudes
and activation statistics \citep{sun2024wanda}. We use Wanda as a baseline in
our structured-pruning experiments. These methods select structure through a
separate scoring procedure. In contrast, \method is a learning method: it
optimizes structural count jointly with the model parameters. It can be applied
to a pretrained model for pruning, but is not limited to post-training use.

\paragraph{Overcomplete learned sparsification.}
Learned sparsification allocates a fixed, overcomplete candidate model and learns
which components survive. Hard-concrete $L_0$ regularization learns a stochastic
gate per parameter or group \citep{louizos2018l0}, while SoRA learns one sparse
gate per LoRA direction \citep{ding2023sora}. Both require a finite candidate set,
and count emerges from many survival decisions. \method instead represents count
directly with a shared boundary over an ordered sequence. Its formulation does
not impose a maximum budget: new components can be instantiated as the boundary
advances.

\paragraph{Flexible computation.}
Elastic models support multiple deployment sizes: DyLoRA, for example, samples
truncation ranks during training so that many LoRA ranks remain usable
\citep{valipour2023dylora}. Adaptive methods instead vary computation by input;
Adaptive Computation Time learns an input-dependent runtime depth
\citep{graves2016act}. \method has a different objective: it learns a fixed
structural allocation for deployment.

\paragraph{Joint structure learning and growth.}
UDN defines an infinite sequence of layers and performs dynamic variational
inference over a global truncation variable \citep{nazaret2022udn}. Each current
posterior has finite support, but the variational family can move to arbitrarily
large depths and instantiate new layers, while prediction averages over supported
depths. AWNN learns unbounded layer width through a decreasing importance
distribution and derives a width from its quantiles \citep{errica2026awnn}.
These methods are closest to our goal of learning capacity during training. UDN
represents a distribution over depth and AWNN derives width from a learned decay.

\section{NGN: Differentiable Structural Count}
\label{sec:method}

\subsection{Differentiable count}

Let \(f_k\) be the \(k\)th member of an ordered sequence of functions and let \(t\)
denote the boundary of the active prefix. During training,
\begin{equation}
  y(x;t)=\sum_{k=0}^{\infty} w_k g_k(t)f_k(x).
  \label{eq:model}
\end{equation}
The relaxed gate is
\begin{equation}
  g_k(t)=\sigmoid\!\left(\beta(t-k+\delta)\right),
  \qquad t=\softplus(\tau)>0.
  \label{eq:gate}
\end{equation}
We use \(\delta=0.5\). As \(\beta\) increases, the gates approach a step
function. Assuming the magnitude of \(f_k(x)\) grows subexponentially with \(k\),
the gated tail vanishes. Under our zero-indexed convention, the retained count is
\(\operatorname{round}(t)+1\), and the deployable model is the corresponding
hard prefix
\begin{equation}
  y_{\mathrm{hard}}(x;t)=
  \sum_{k=0}^{\operatorname{round}(t)}w_k f_k(x).
  \label{eq:hard}
\end{equation}
Throughout, \emph{soft} evaluation uses the whole candidate model in
Equation~\ref{eq:model} with its learned sigmoid gates, whereas \emph{truncation}
uses the hard prefix in Equation~\ref{eq:hard}. We report the optimized boundary
\(t\) and, where relevant, the retained count separately; thus, \(t\)
is a continuous proxy for the discrete structural count rather than the count
itself.

We optimize the task loss together with a capacity price and an integer penalty,
\begin{equation}
  \mathcal J
  =\mathcal L_{\mathrm{task}}
  +\lambda t
  +\alpha\sin^2(\pi t).
  \label{eq:objective}
\end{equation}
The term \(\lambda t\) penalizes the size parameter, and
\(\alpha\sin^2(\pi t)\) has minima at the integers, so it snaps the learned
boundary to an integer. During training, a nonzero \(\lambda\) can shift the
minimum of the combined regularizer away from an integer. We therefore anneal
\(\lambda\) to zero in the standard schedule, leaving the periodic term with
exact integer minima at the end of training.

The gradient with respect to \(t\) explains how the boundary moves:
\begin{align}
  \frac{\partial y}{\partial t}
   & =\sum_k w_k\beta g_k(t)(1-g_k(t))f_k(x),
  \label{eq:dydt}                             \\
  \frac{\partial\mathcal J}{\partial t}
   & =\left\langle
  \nabla_y\mathcal L_{\mathrm{task}},
  \frac{\partial y}{\partial t}\right\rangle
  +\lambda+\alpha\pi\sin(2\pi t).
  \label{eq:djdt}
\end{align}
Because \(g_k(1-g_k)\) is concentrated at the boundary, the task gradient is
dominated by the current frontier. A useful boundary function supplies enough
recruiting force to move \(t\) outward; otherwise the capacity price moves it
inward.

This mechanism requires the ordered prefix to be genuinely capacity-increasing.
If \(\mathcal H_K\) is the hypothesis class represented by the first \(K\) functions,
we require \(\mathcal H_K\subsetneq\mathcal H_{K+1}\). Thus, for some tasks,
adding \(f_K\) permits a strictly better fit. However, the \(f_k\) need not have
identical architectures.

\subsection{Implementation and training schedules}

For simplicity, our experiments instantiate \(K\) candidate functions before
training. An equivalent but more involved implementation could dynamically
allocate new trainable weights as \(t\) grows, yielding an unbounded model without
a preset \(K\); we do not use it here.

We set \(w_k=2^{-k}\) to order the candidates and attenuate later functions.
Section~\ref{sec:learned-size} shows that this ordering also improves the
stability of the learned boundary.

Annealing the gate sharpness \(\beta\) is essential for faithful truncation: a
soft gate early in training lets gradients reach the frontier, while a sharp gate
late in training approaches the hard prefix. A simple linear increase such as
\(4\!\rightarrow\!12\) works well in our experiments.

We zero-initialize \(f_k\) as a convenient default; the fixed ordering already
breaks symmetry among functions, so distinct initialization scales are unnecessary.

% Keep the experimental results in their source file so updates to data.tex are
% reflected here without duplicating them.
\input{data.tex}

\section{Limitations and future directions}

The experiments demonstrate broad transfer across structures, but they also
reveal several limitations on when count learning is effective.

\paragraph{Ordered candidates.}
\method{} turns a structural hyperparameter into an optimized scalar, but it does
not make arbitrary architecture choices differentiable.  Candidates must form an
ordered, nested sequence in which each function extends the current prefix.  The
boundary cannot skip a harmful early function to reach a useful later one.

\paragraph{Finite growth range.}
Geometric weights order the candidates and suppress the soft tail, but also limit
growth.  Because $w_k=0.5^k$ decays exponentially, high-index functions receive
coefficients too small to recruit reliably in our experiments.  The block size
must therefore be chosen so that the plausible capacity range can be represented
with a moderate number of functions.  One way to bypass this limit is to learn
capacity across multiple layers, although the learnable width within each layer
remains limited.  Ordering schemes without exponential attenuation are an
important direction for future work.

\paragraph{Optimization cost.}
In our experiments, \method{} takes more optimization steps to reach the same
performance as a fixed architecture because it learns the boundary and function
parameters jointly while annealing a soft frontier. With the same step budget, \method{} is
often similar to or slightly worse than a fixed model.  Its benefit is automatic
size selection, not faster optimization or consistently better performance at the
selected size.

\paragraph{Hyperparameter sensitivity.}
\method{} does not eliminate hyperparameter tuning.  In particular, the capacity
price $\lambda$ must be calibrated to obtain the desired size--performance
trade-off, while insufficient effective snapping pressure from $\alpha$ can leave
the boundary away from an integer and weaken hard truncation.  New losses or
function scales may therefore require adjusting both $\alpha$ and the
gate-sharpening schedule.  The large-model PEFT results illustrate that these
effects are distinct: integer rates can remain low on certain tasks, consistent with insufficient
snapping pressure, while the annealed $\beta$ schedule still keeps many soft and
truncated scores close (Appendix Table~\ref{tab:data11-diagnostics}).  Changing
$\lambda$ can retarget a converged model
(Appendix~\ref{app:pruning-1d}), but the resulting architecture must still be
validated.

\paragraph{Interpreting learned allocations.}
Future work should develop interpretability analyses of learned shapes to explain
where a network uses capacity.  In Table~\ref{tab:data11}, for example, the
learned LoRA ranks for both WikiSQL variants concentrate in the middle layer
third.  Testing whether such patterns persist across tasks, seeds, and models may
reveal distinct roles for different layers.

\section{Conclusion}

We presented \method{}, a direct differentiable representation of how many
ordered components a neural network should use.  One boundary variable controls
a nested prefix; a capacity price moves that boundary, a periodic penalty resolves
it to an integer, and gate sharpening makes the learned soft model deployable by
truncation.  Controlled experiments separate these roles and show why learning a
count is different from learning a normalized decay rate.  Across feature functions,
convolution and graph branches, Transformer components, residual depth, LoRA,
and adapters, the same primitive learns heterogeneous structural
budgets without changing its interpretation.  The results also identify the
boundary of the claim: count learning is most useful when capacity is naturally
ordered, and the learned size remains tied to the task objective and its chosen
price.

\subsection*{Reproducibility statement}
Sections~\ref{sec:method}--\ref{sec:experiments} provide the model and experiments.
Appendices~\ref{app:full-uncertainty}--\ref{app:settings} report full uncertainty
results, the synthetic manifold construction, architectures, structural settings,
hardware, and training-time ranges. Code and configuration files will be included
in the supplementary material.

\subsection*{AI use statement}
Generative AI tools were used to aid and polish writing, retrieve and discover
related work, support research ideation and execution, and draft portions of the
manuscript. The authors independently checked cited sources, validated
AI-assisted research and experimental work, and reviewed all AI-generated text.
The authors take responsibility for the final content of this work.

\bibliography{main}
\bibliographystyle{\paperbibliographystyle}

\appendix

\UncertaintyTablesAppendix

\ManifoldSetupAppendix

\ExperimentalSettingsAppendix

\DecayAblationAppendix

\AdaptiveRankBaselinesAppendix

\DatasetLicensesAppendix

\PruningAblationAppendix

\ContinueLearningAppendix

%% file: data.tex
% This file contains all the data used in the paper
% Numbers/tables below come from build_paper_data.py -> data_generated.tex
\input{data_generated.tex}

\section{Controlled analysis}
\label{sec:analysis}

We use a controlled one-dimensional regression task to isolate the behavior of
the count gate. Figure~\ref{fig:data1} defines the NGN baseline used throughout
this section. Architectures and structural
hyperparameters for all Section~\ref{sec:analysis} studies are collected in
Appendix~\ref{app:settings}, Table~\ref{tab:settings-analysis}.

\beginpaperfigure
  \centering
  \includegraphics[width=\textwidth]{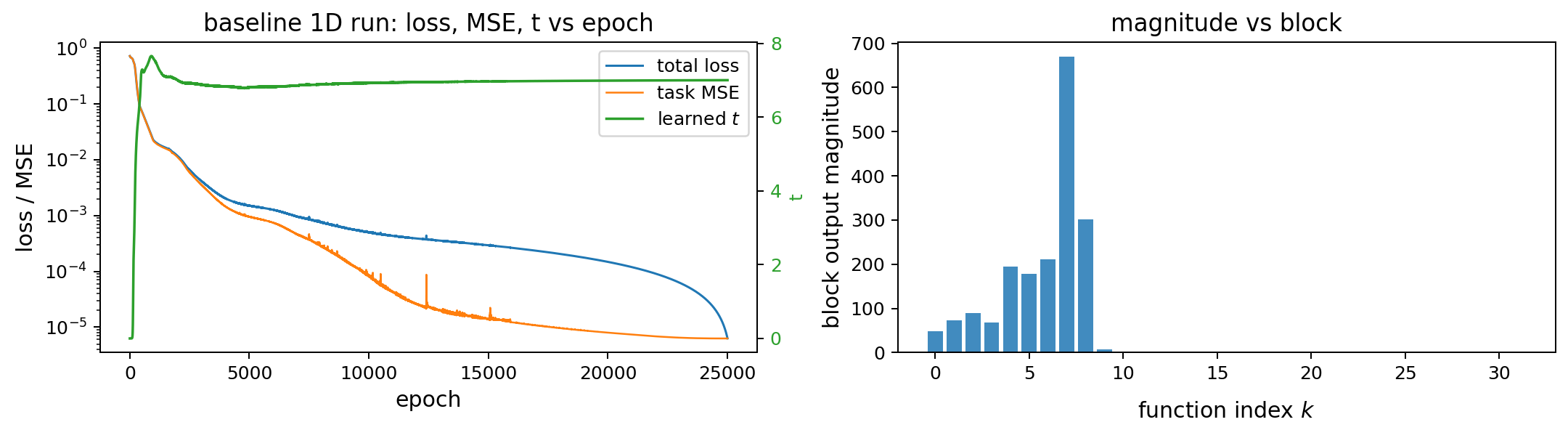}
  \caption{Baseline setting for the controlled 1D regression. The target is
  $y^*(x)=\sin(3x)+0.6\sin(7x)+0.3\sin(13x)$ on $[-\pi,\pi]$, and each
  $f_k$ is a tanh MLP. Unless stated otherwise, the NGN baseline uses
  $w_k=0.5^k$, $\lambda_0=\alpha_{\max}=10^{-4}$, linearly decreasing $\lambda$,
  increasing $\alpha$, and $\beta:4\!\to\!15$. Left: total loss, task MSE,
  and learned $t$ during training. Right: $\|f_k\|_2$ for each function.
  Final $t=\stdT$, soft MSE $=\stdSoftMSE$, and truncation MSE
  $=\stdHardMSE$.}
  \label{fig:data1}
\end{figure}

\subsection{Hyperparameter sensitivity and schedule ablations}

Figure~\ref{fig:data2} shows that $\lambda$ is a practical capacity knob:
increasing it generally trades task fit for a smaller count, while integer
snapping produces a staircase of stable count plateaus. Away from the
irregular middle transition, most plateaus span roughly a factor of two in
$\lambda$ on the log scale, so the selected count is robust to modest
$\lambda$ changes.

\beginpaperfigure
  \centering
  \includegraphics[width=0.78\textwidth]{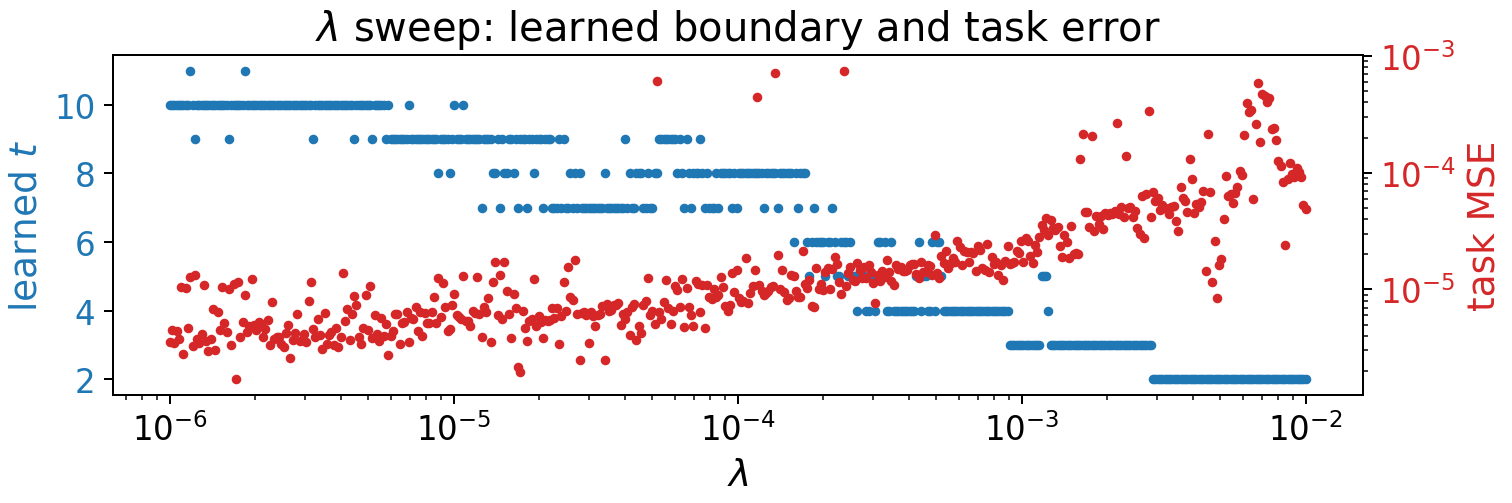}
  \caption{$\lambda$ sweep from $10^{-6}$ to $10^{-2}$ over 500 models, with
    the other NGN settings fixed to the baseline. Increasing $\lambda$ trades
    task fit for a smaller count. The learned boundary forms integer plateaus,
    showing that snapping remains effective across the sweep. }
  \label{fig:data2}
\end{figure}

\beginpapertable
  \centering
  \caption{Schedule ablation on the 1D task (10 seeds per setting; medians).
    The upper block tests robustness to schedule shape; the lower block removes
    or freezes components involved in truncation or growth. ``Integer''
    is the percentage of final boundaries within $0.01$ of an integer. Full
    mean$\,\pm\,$standard deviation results are in
    Table~\ref{tab:data4-full}.}
  \label{tab:data4}
  \begin{tabular}{lcccc}
    \toprule
    setting & $t$ & integer & soft MSE & truncated MSE \\
    \midrule
    \datafourbody
    \bottomrule
  \end{tabular}
\end{table}

Table~\ref{tab:data4} separates robustness to schedule shape from the observed
roles of its components on this controlled task. Constant $\lambda$ and constant
$\alpha$ yield comparable median task fit in grow-from-low training, so the
behavior does not depend on one exact schedule choice. We retain decreasing
$\lambda$ and increasing $\alpha$ because Figure~\ref{fig:data5} shows that
delayed $\alpha$ improves prune-from-high recovery, while annealing $\lambda$ to
zero removes the linear capacity price at the endpoint and leaves the periodic
term minimized at integers. In these ablations, removing $\alpha$ or holding
$\beta=4$ increases the soft-to-truncated error gap, whereas starting at
$\beta=15$ selects smaller counts with worse median fit. These observations
support distinct roles for snapping and gate sharpening in this setting.

\subsection{Capacity-matched comparisons}
\label{sec:capacity-comparison}

Fixed prefixes trained from scratch trace nearly the same error--count frontier
as grown NGN models and are typically slightly better at matched retained count
(Figure~\ref{fig:data3}). The benefit here is automatic size selection.

\beginpaperfigure
  \centering
  \includegraphics[width=0.62\textwidth]{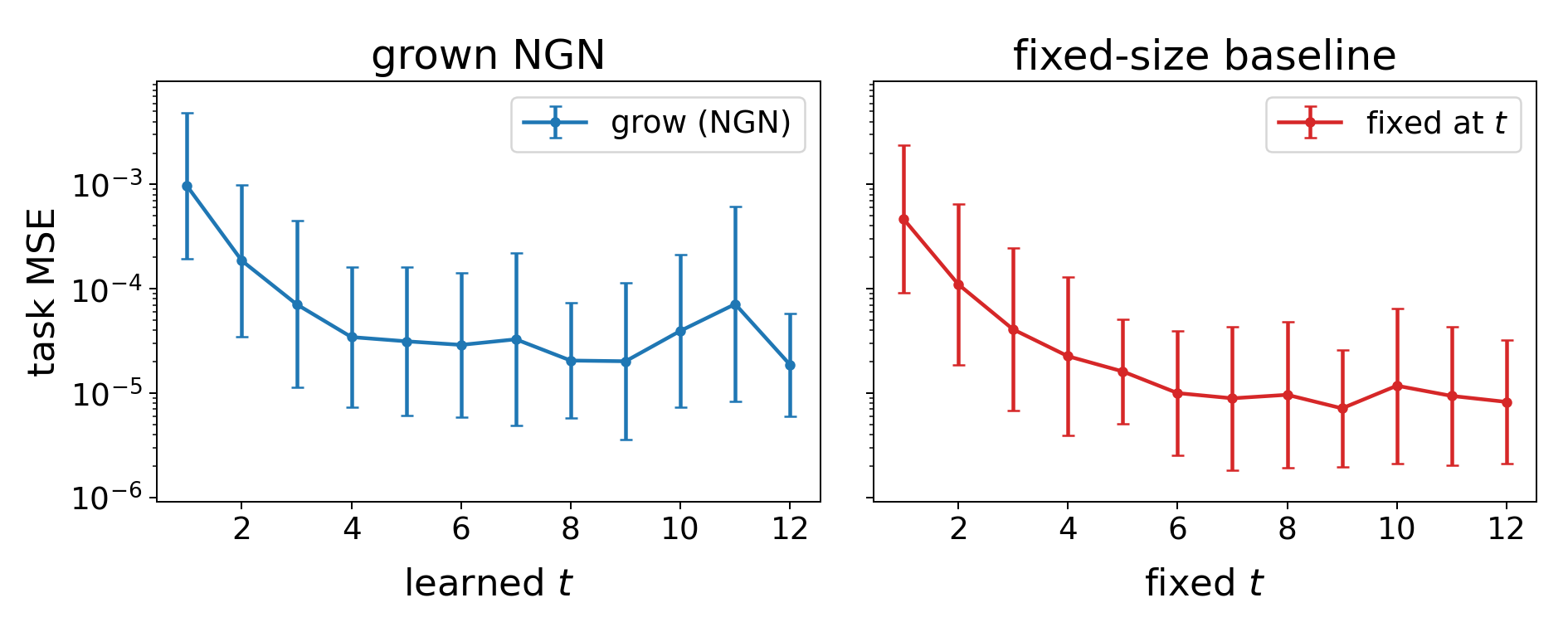}
  \caption{Grown NGN models and fixed-size prefixes trained from scratch at
    matched retained counts over 50 seeds. Lines show geometric mean MSE, and asymmetric
    error bars show one standard deviation in $\log_{10}(\mathrm{MSE})$. The
    two frontiers are similar, with fixed models typically slightly better at
    the same count.}
  \label{fig:data3}
\end{figure}

We next compare three learned-capacity baselines. AWNN follows the learned
importance distribution and quantile-width rule of \citet{errica2026awnn},
applied to the same candidate MLP blocks so all methods select the same unit of
count. $L_0$ uses one independent hard-concrete gate per block with the
expected-$L_0$ regularizer of \citet{louizos2018l0}. STE keeps NGN's single
boundary and objective but uses the hard prefix in the forward pass and the
sigmoid derivative in the backward pass.

Appendix~\ref{app:decay-ablation} investigates possible contributors to the
underperformance of this controlled AWNN implementation by separately testing
its learned decay rate, normalization, and quantile-derived width in the common
NGN function bank.

Within the overlapping deployed-count range, $L_0$ generally occupies a
higher-error region than NGN (Figure~\ref{fig:data7}). AWNN selects a more stable
count across seeds but fits worse. STE can approach NGN when growth succeeds, yet
stalls at small counts more often. This pattern is consistent with hard forward
gating, under which inactive frontier blocks receive no body-weight updates until
they activate; NGN's soft gate supplies such gradients before hard truncation at
deployment.

\beginpaperfigure
  \centering
  \includegraphics[width=0.82\textwidth]{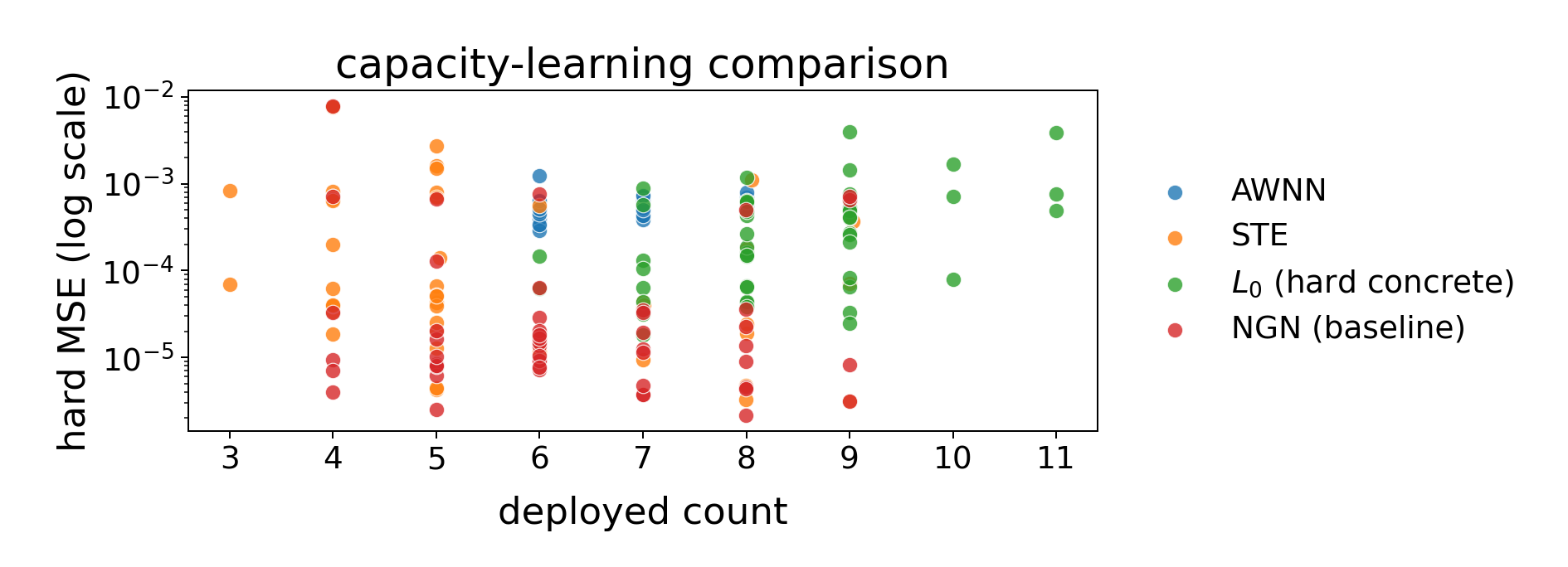}
  \caption{Capacity-learning comparison over 50 seeds and 25,000 epochs per
    seed. Each point is one seed's deployed MSE against retained count.}
  \label{fig:data7}
\end{figure}

\subsection{Stability of learned size}
\label{sec:learned-size}

Pruning is desirable because growth can overshoot the useful count, as the
training trace in Figure~\ref{fig:data1} illustrates. A stable count learner
should therefore move in both directions. Here we test that property in the
controlled 1D setting; Section~\ref{sec:experiments} checks prune-from-high
behavior on deeper models, where it is not universally observed. With geometric weights ($r=0.5$) and
delayed $\alpha$, grow-from-low and prune-from-high runs converge to a similar
final range (Figure~\ref{fig:data5}). Uniform weights and applying $\alpha$ at
full strength from the start both fail to prune reliably. This also explains why
the baseline delays $\alpha$ even though constant $\alpha$ can fit the
grow-from-low task.

\beginpaperfigure
  \centering
  \includegraphics[width=\textwidth]{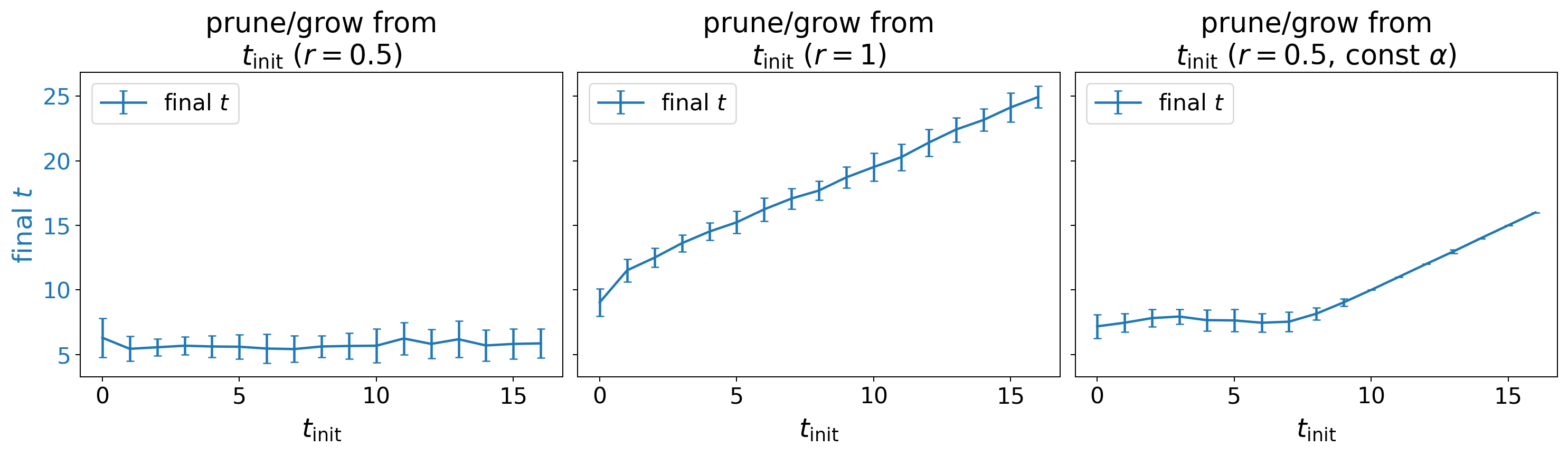}
  \caption{Grow/prune stability over $t_\text{init}\in[0,16]$, with 50 seeds
    per initialization. Lines show mean final $t$, and error bars indicate one
    standard deviation. With geometric weights ($r=0.5$, left) and delayed $\alpha$,
    grow and prune initializations converge to a similar range. Uniform weights
    ($r=1$, middle) and constant $\alpha$ from the start (right) fail to prune
    reliably.}
  \label{fig:data5}
\end{figure}

\subsection{Analysis of Intrinsic Capacity Learning}
\label{sec:manifold-dimension}

We next ask whether NGN's selected capacity tracks a known structural target. We
study autoencoding on a nonlinear manifold whose intrinsic dimension $D$ is a
natural target bottleneck size: exact continuous reconstruction must preserve
its $D$ degrees of freedom. We use a variational autoencoder (VAE)
\citep{kingma2014vae} for this task. Each $f_k$ contributes one rank-one bottleneck term,
so recruiting $D$ functions provides capacity for $D$ independent dimensions.
Across 50 seed--dimension cases, the selected count and numerical bottleneck
rank match the intrinsic dimension in 49 (98\%). Thus NGN recovers the target
capacity in nearly every run, while the remaining error and occasional
soft--truncated gap quantify optimization variability.
The construction, architecture, and training details
are in Appendix~\ref{app:manifold-setup}.

\beginpapertable
  \centering
  \small
  \caption{Capacity tracking across nonlinear manifolds embedded in the
    16-dimensional ambient space, over ten optimization seeds (medians; full
    mean and standard deviation results are in
    Table~\ref{tab:data15-full}). ``Learned $D$'' is the selected number of
    rank-one bottleneck terms after truncation, and ``exact rate'' is the fraction of seeds for
    which learned $D$ equals the true intrinsic dimension. Normalized mean squared error is
    \(
    \mathrm{NMSE}=
    \mathbb{E}\lVert\hat{x}-x\rVert_2^2 /
    \mathbb{E}\lVert x\rVert_2^2.
    \)}
  \label{tab:data15}
  \begin{tabular}{rcccc}
    \toprule
    true $D$ & learned $D$ & exact rate & soft NMSE & trunc. NMSE \\
    \midrule
    \datafifteenbody
    \bottomrule
  \end{tabular}
\end{table}

\section{Generalization across structures and tasks}
\label{sec:experiments}

The controlled study isolates the behavior of one boundary within a common
function bank. We next test whether the same boundary parameterization transfers
when the ordered components represent different structures and when several
boundaries are learned within one model.
Appendix~\ref{app:settings} gives the complete block formulas and Table~\ref{tab:settings-experiments}
lists every structural schedule. Dataset sources and licenses are recorded in
Appendix~\ref{app:dataset-licenses}.

\subsection{Ordered components across tasks}

\beginpapertable
  \centering
  \small
  \caption{NGN applied across architectures and tasks. Scores are
    test accuracy (\%, higher is better) or TinyStories validation bits/character
    (lower is better). The four non-language-model rows report medians over ten
    seeds; the TinyStories rows remain single-seed experiments. ``Fixed''
    retrains the hard architecture at each seed's learned size; the final column
    starts from high capacity and reports whether pruning returns to the
    grow-from-low allocation. Full mean$\,\pm\,$standard deviation results for
    the ten-seed rows and detailed prune-from-high outcomes are in
    Table~\ref{tab:data8-full}. For each seed, ``yes'' means that every boundary
    returns to within one block of its grow-from-low allocation, ``partial''
    means that every boundary moves more than one block below its high-capacity
    initialization without satisfying that
    return criterion, and ``no'' covers the remaining cases. The aggregate row
    is ``yes'' or ``no'' only when every seed has that outcome and ``partial''
    otherwise. ``Integer'' is the percentage of final learned boundaries within
    $0.01$ of an integer.}
  \label{tab:data8}
  \setlength{\tabcolsep}{4pt}
  \resizebox{\textwidth}{!}{%
    \begin{tabular}{llcccccl}
      \toprule
      task & $f_k$ & learned $t$ & integer & soft & trunc & fixed & prunes back? \\
      \midrule
      \dataeightbody
      \bottomrule
    \end{tabular}}
\end{table}

Across these tasks, \method{} learns useful counts for MLP, convolutional
\citep{lecun1998mnist}, graph \citep{kipf2017gcn}, Transformer
\citep{vaswani2017attention}, and Mamba \citep{gu2023mamba} components,
including several gates across multiple layers at once. Hard truncation usually
costs little and the selected
architectures perform similarly to matched fixed retraining, showing that count
learning transfers across losses and function types. Because reliable
prune-from-high recovery is not observed across the full set, the breadth result
supports growth and truncation only.

\subsection{Learning depth}

We next ask whether the count boundary can learn model depth. We test two constructions. First, a multi-layer model learns the
width at each layer with a separate boundary $t_\ell$; for
the removable layers, $t_\ell=0$ retains only $f_{\ell 0}=I$, the
identity map, and the model learns its effective depth. Second, a recursive model
uses one global boundary over functions $f_k$ with $k+1$ composed layers, so its
selected prefix directly determines the length of the computation. Complete
definitions are given in Appendix~\ref{app:settings},
Equations~\ref{eq:depth-per-layer}--\ref{eq:depth-recursive}.

\beginpapertable
  \centering
  \small
  \caption{Four-setting sweep and a recursive count construction for depth on
    the iterated 2D twist task. Per-layer rows learn one boundary $t_\ell$ at each
    of ten residual layers; the recursive row learns one global boundary over a
    chain of at most sixteen functions. For the per-layer construction,
    const-$r$ fixes $r_\ell=0.5$, whereas inc-$r$ increases $r_\ell$ linearly from
    $0.5$ in the first layer to $0.75$ in the last. Const-$\lambda$ uses the
    same capacity price at every layer, while dec-$\lambda$ decreases its
    initial value geometrically across layers from $10^{-1}$ to $10^{-5}$;
    every layer then follows the temporal power-decay schedule in Appendix
    Table~\ref{tab:settings-experiments}. Learned
    depth counts non-identity layers in the per-layer construction and retained
    composed functions in the recursive construction.}
  \label{tab:data9}
  \resizebox{\textwidth}{!}{%
    \begin{tabular}{lccccc}
      \toprule
      construction / setting & learned $t$ & depth & soft MSE & trunc. MSE & fixed MSE \\
      \midrule
      \dataninebody
      \datatenbody
      \bottomrule
    \end{tabular}}
\end{table}

For the per-layer construction, clean depth learning appears only when we use
both proposed schedules: geometrically decreasing $\lambda$ produces a
contiguous identity prefix, while increasing $r$ across layers preserves a
well-fitting ordered active suffix of depth six. Constant $\lambda$ leaves all
ten layers active, and geometric-decay $\lambda$ with fixed $r$ learns depth
seven but gives a substantially worse fit. The recursive construction works
without a depth sweep, learning depth eight ($t=7.02$) with closely matched soft
and truncated errors. Thus NGN
can learn sequential depth either through scheduled per-layer widths or through
one boundary over recursively deepening functions.

\subsection{Learning LoRA rank and adapter width}

% data 11: table of scores on different tasks
% basemodel is llama3-8b-instruct, pick the lastest record from results.json
% column, basemodel score, lora total rank, e/m/l rank (this is one column), soft/hard score, fixed baseline, ppl on forgetting/basemodel, fixed ppl/basemodel
% column(write below lora), basemodel score, adapter total width, e/m/l width (this is one column), soft/hard score, fixed baseline, ppl on forgetting/basemodel, fixed ppl/basemodel
% row (copied twice for lora and adapter), wikisql-easy, wikisql-hard, spider, gsm8k, squad (use 20ep llama for wikisql)
% e/m/l is early 1/3, mid 1/3, late/13
\beginpapertable
  \centering
  \small
  \caption{Learned PEFT capacity on Llama-3-8B-Instruct (single seed per
    configuration). NGN reports a soft/truncated task score and a fixed model
    trained with the same rounded per-layer allocation; parentheses split NGN
    size across early/middle/late layer thirds.
    SoRA and AdaLoRA target NGN's learned LoRA rank, but SoRA reaches a lower
    rank on Spider and GSM8K. NGN scores higher than SoRA on the three exact
    matches and higher than AdaLoRA on four of five tasks. WikiSQL-easy has one
    \textsc{where} condition; WikiSQL-hard combines all multi-condition queries
    with an equally sized seeded sample of single-condition queries. All four
    LoRA methods use 10 epochs on WikiSQL; the WikiSQL adapter runs use 20 epochs.
    Integer and forgetting diagnostics are in
    Appendix~\ref{app:adaptive-rank-baselines}.}
  \label{tab:data11}
  \setlength{\tabcolsep}{4pt}
  \begin{tabular}{lcccccc}
    \toprule
    \multicolumn{7}{c}{\textbf{LoRA rank}}                                    \\
    \midrule
    task & base & NGN rank (e/m/l) & NGN soft/trunc. & fixed & SoRA & AdaLoRA \\
    \midrule
    \dataelevenlorabody
    \bottomrule
  \end{tabular}

  \vspace{4pt}
  \begin{tabular}{lcccc}
    \toprule
    \multicolumn{5}{c}{\textbf{Adapter width}}                \\
    \midrule
    task & base & NGN width (e/m/l) & NGN soft/trunc. & fixed \\
    \midrule
    \dataelevenadapterbody
    \bottomrule
  \end{tabular}
\end{table}

NGN learns heterogeneous PEFT budgets while staying close to its hard
truncation and matched fixed models. SoRA learns one sparse gate per LoRA
direction using proximal $\ell_1$ optimization \citep{ding2023sora}, placing it in
the same broad independent-gate family as the $L_0$ baseline in
Section~\ref{sec:capacity-comparison}. AdaLoRA scores and prunes singular triplets
toward a prescribed rank budget \citep{zhang2023adalora}.
For both PEFT constructions we use the shifted gate $\delta=-0.5$. Thus $t=0$
retains zero LoRA directions or adapter blocks, while hard truncation retains
exactly $\operatorname{round}(t)$ blocks in that layer.

Both baselines instantiate a finite maximum candidate rank, and AdaLoRA also
requires the final target rank. NGN instead optimizes count directly with one
ordered boundary per layer, a representation that can allocate new functions as
the boundary advances. The adapter rows show that the same mechanism transfers
from linear rank to nonlinear bottleneck width.

\newcommand{\ContinueLearningAppendix}{%
  \subsection{1D pruning and post-hoc retargeting}
  \label{app:pruning-1d}
  A poor initial choice of \(\lambda_0\) need not require restarting the model.
  After the standard run converges, we restart the schedules and continue from the
  same checkpoint with the capacity price multiplied by \(0.1\), \(1\), or \(10\).
  The shared trajectory then separates according to the new price
  (Figure~\ref{fig:data6}): the $10\times$ price prunes strongly, the unchanged
  price prunes modestly, and the $0.1\times$ price preserves the original
  structural plateau. Thus increasing the price can remove capacity after the
  function parameters have learned the task, whereas lowering it does not by
  itself recruit a new component once the task is already fit.
  % this is a study on whether the model can change size after done training
  % run and cache a model after a standard training, do another standard run on the model only changing lambda
  % data 6: single plot of t and mse vs epoch (conbine step 1 and step 2 training in the same plot like current)
  % the plot should bifercate after step 1 phase into 3 lambda = 10*, lambda = 1*, lambda = 0.1* the standard setting
  % Figure 6 was removed; retain the established numbering of later figures.
  \begin{figure}[H]
    \centering
    \includegraphics[width=0.7\textwidth]{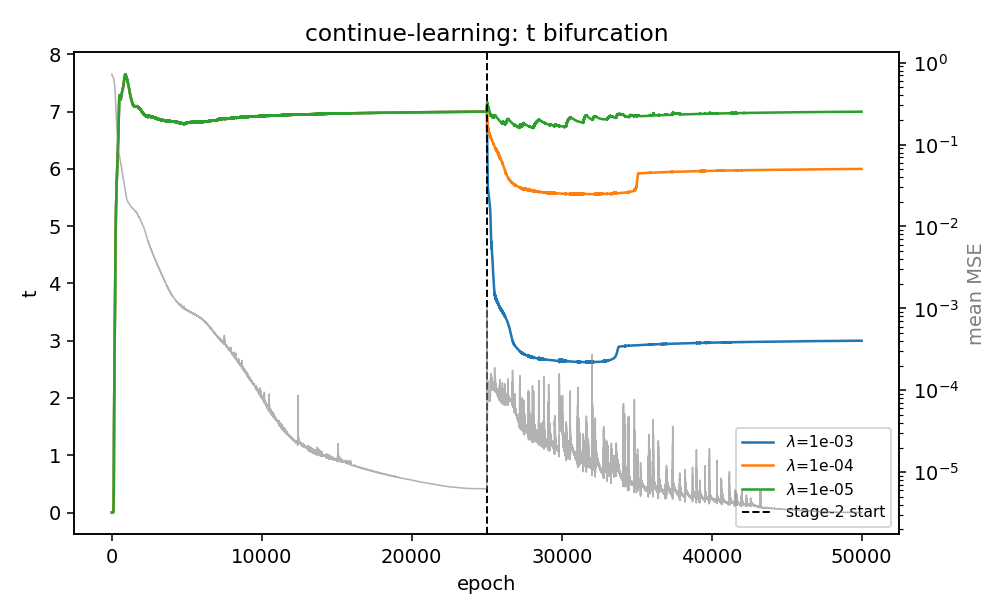}
    \caption{Continue-learning. A model is trained to convergence (stage 1, shared),
      then training continues with only $\lambda$ changed. After the stage-2 boundary (dashed),
      larger $\lambda$ produces stronger pruning, while $0.1\times$ the standard
      price preserves the original plateau rather than recruiting new capacity.}
    \label{fig:data6}
  \end{figure}
}

% Pruning-price ablation, invoked after \appendix by the paper.
\newcommand{\PruningAblationAppendix}{%
\section{Pruning frontier and settings}
\label{app:pruning-ablation}

\subsection{Direct structured pruning}

\begin{table}[H]
  \centering
  \small
  \caption{Direct FFN pruning of Llama-3-8B-Instruct on WikiSQL-easy.
    NGN learns the kept dimension; random removal is a matched-budget control, and
    Wanda ranks weights using weight magnitude and activation statistics
    \citep{sun2024wanda}. ``Mixed'' adds general-domain replay data to the task
    data to reduce catastrophic forgetting. Task-only and mixed columns use
    matched budgets; random and Wanda reuse NGN's per-layer allocation and
    differ only in which neurons they retain within each layer. ``Kept'' is the
    percentage of the original FFN intermediate dimensions retained. The e/m/l
    entry gives the percentage retained within each of the early, middle, and
    late layer thirds. Forgetting is Pile perplexity relative to
    the base model. Among NGN's final
    per-layer boundaries, \datafourteenintrate{} (task-only) and
    \datafourteenmixintrate{} (mixed) land within $0.01$ of an integer.}
  \label{tab:data14}
  \setlength{\tabcolsep}{5pt}
  \begin{tabular}{lcccccccc}
    \toprule
           & \multicolumn{4}{c}{task-only} & \multicolumn{4}{c}{mixed}                                                                   \\
    \cmidrule(lr){2-5}\cmidrule(lr){6-9}
    method & kept                          & e/m/l (\%)                & score & ppl $\times$ & kept & e/m/l (\%) & score & ppl $\times$ \\
    \midrule
    \datafourteenbody
    \bottomrule
  \end{tabular}
\end{table}

The same count variable can therefore learn a direct structured-pruning
budget. NGN is far stronger than random removal and broadly competitive with
Wanda \citep{sun2024wanda}; its distinct advantage is not a stronger
within-budget importance score, but that the kept dimension and allocation are
learned jointly from the task and replay objective instead of being supplied
before pruning. The following studies report the full
capacity--accuracy--forgetting frontier and its structural settings.

\subsection{Pythia-410M pruning frontier}
Figure~\ref{fig:data13} and Table~\ref{tab:data13} vary the capacity price and
the gated structure for Pythia-410M on WikiSQL-easy. The resulting ladder traces
a monotone practical frontier: lowering the price retains more weight, raises
exact match, and reduces forgetting. Freezing the surviving backbone weights is
substantially safer than updating the full model at comparable target score.
\begin{figure}[H]
  \centering
  \includegraphics[width=\textwidth]{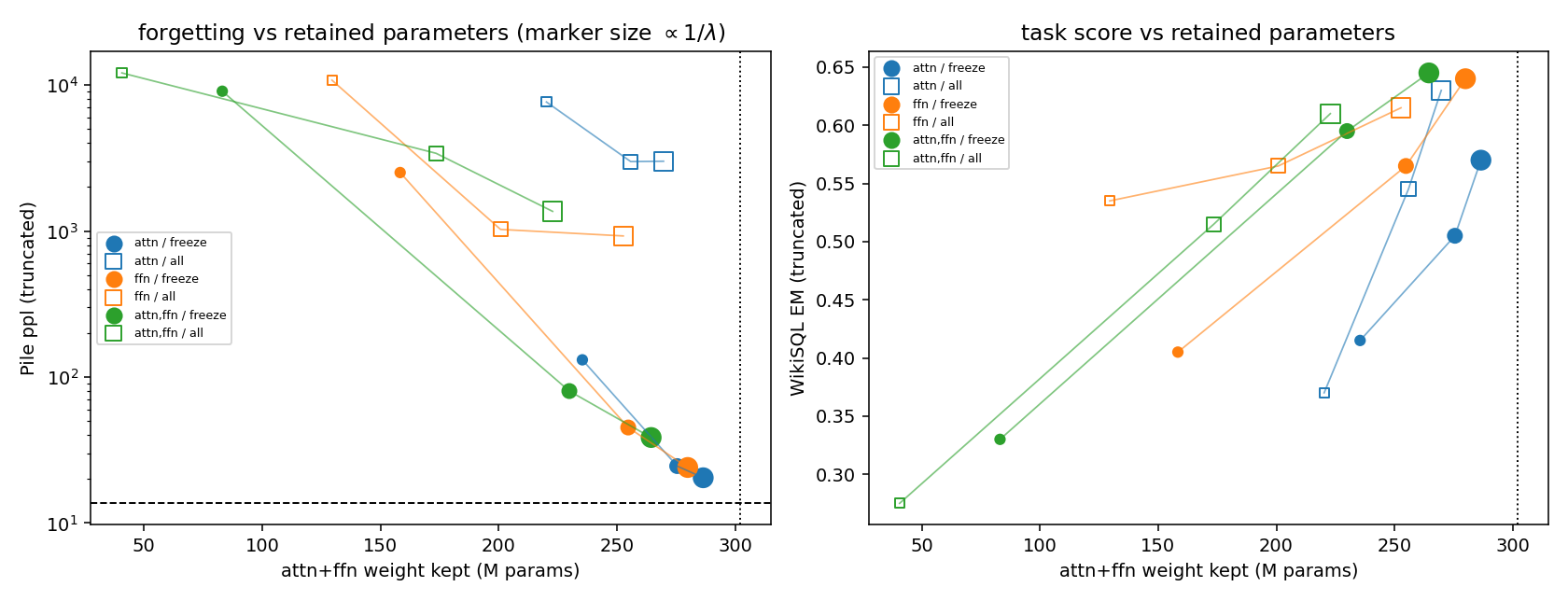}
  \caption{Structured-pruning hyperparameter ablation for Pythia-410M on
    WikiSQL-easy. Marker size increases as $\lambda$ decreases; filled markers
    freeze surviving weights and open markers update the full model.}
  \label{fig:data13}
\end{figure}

\subsection{Pythia-410M pruning settings}
The gate partitions either attention heads, FFN neurons, or both in
Pythia-410M \citep{biderman2023pythia} into 16 ordered blocks per layer. Here
$f_k$ is the contribution of the $k$th retained structural
block, $t$ starts from 15, $w_k=1$, $\alpha_{\max}=0.03$, and
$\beta:4\!\to\!12$; the sweep varies $\lambda\in\{10^{-2},10^{-3},10^{-4}\}$
with a constant capacity-price schedule.

\begin{table}[H]
  \centering
  \small
  \caption{Numerical results for all frozen- and updated-backbone runs in the
    pruning hyperparameter ablation of Figure~\ref{fig:data13} (10 epochs).
    ``Kept'' is the percentage
    of the full 302M attention-plus-FFN parameter budget retained, and e/m/l gives
    the retained percentage within each layer third. We also report truncated
    exact match and Pile perplexity relative to the base model.}
  \label{tab:data13}
  \resizebox{\textwidth}{!}{%
    \begin{tabular}{llccccc}
      \toprule
      target & backbone & $\lambda$ & kept & e/m/l (\%) & EM (trunc) & Pile $\times$base \\
      \midrule
      \datathirteenbody
      \bottomrule
    \end{tabular}}
\end{table}

At $\lambda=10^{-4}$, pruning both structures keeps 87.6\% of the
attention-plus-FFN parameters
and reaches 64.5\% exact match, while Pile perplexity remains $2.83\times$ the
base model's. More aggressive joint pruning keeps 27.5\% and raises that multiplier
to $665\times$. This ablation supports learned budget selection for a target
task, not lossless general-purpose compression.
}

\newcommand{\ExperimentalSettingsAppendix}{%
\section{Experimental settings and hyperparameters}
\label{app:settings}
We report the settings that determine the learned structure: the architecture,
the candidate function $f_k$, the candidate count $K$, and
$(\lambda_0,\alpha_{\max},\beta,w_k)$. Unless marked otherwise, $\lambda$
decreases with power one, $\alpha$ begins a linear increase halfway through
training, and $\beta$ increases linearly. Fixed controls use the stated hard
count with no capacity penalty. Optimizer, learning rate, gradient clipping, and
learning-rate schedule details are omitted because they are not varied as
structural hyperparameters. The implementation optimizes the unconstrained
parameter $\tau$ with $t=\softplus(\tau)$. A nominal initialization of $t_0=0$
is represented numerically by $t_0=10^{-4}$ before applying the inverse
softplus.

For the parallel controlled models, the entire learned block is
\begin{equation}
  y(x)=\sum_{k=0}^{K-1}w_k g_k(t)
  \left[W_{3k}\tanh\!\left(W_{2k}\tanh(W_{1k}x+b_{1k})+b_{2k}\right)+b_{3k}\right].
  \label{eq:settings-1d}
\end{equation}
The manifold experiment uses the variational rank-one bottleneck defined in
Appendix~\ref{app:manifold-setup} between the encoder and decoder.

\subsection{Section~\ref{sec:analysis}: controlled studies}
\begin{table}[H]
  \centering
  \small
  \caption{Architectures and structural hyperparameters for the controlled
    studies in Section~\ref{sec:analysis}.}
  \label{tab:settings-analysis}
  \resizebox{\textwidth}{!}{%
    \begin{tabular}{lccccccc}
      \toprule
      study                & $K$ & $\lambda_0$           & $\lambda$ sched.      & $\alpha_{\max}$ & $\alpha$ sched.       & $\beta$               & $w_k$           \\
      \midrule
      1D standard          & 32  & $10^{-4}$             & power-decay           & $10^{-4}$       & delayed linear        & $4\!\to\!15$          & $0.5^k$         \\
      1D $\lambda$ sweep   & 32  & $10^{-6}\!:\!10^{-2}$ & power-decay           & $\lambda_0$     & delayed linear        & $4\!\to\!15$          & $0.5^k$         \\
      initialization sweep & 32  & $10^{-4}$             & power-decay           & $10^{-4}$       & delayed linear        & $4\!\to\!15$          & $r^k$           \\
      schedule ablations   & 32  & $10^{-4}$             & Table~\ref{tab:data4} & $10^{-4}$       & Table~\ref{tab:data4} & Table~\ref{tab:data4} & $0.5^k$         \\
      capacity baselines   & 32  & selected              & method-specific       & method-specific & method-specific       & method-specific       & method-specific \\
      manifold rank        & 10  & $2\!\times\!10^{-3}$  & linear decay          & $10^{-3}$       & delayed linear        & $4\!\to\!12$          & $0.5^k$         \\
      \bottomrule
    \end{tabular}}
\end{table}

For the capacity-learning comparison, all methods use the same 32 candidate
MLP blocks, 512 training samples, 25,000 optimization steps, body learning
rate $10^{-3}$, and gradient clipping at $0.003$. Configurations were chosen
to match NGN's mean retained count of eight blocks. AWNN uses initial width
16, 0.9 quantile mass, Gaussian-prior standard deviation 10, and controller
learning rate $8\times10^{-5}$. Hard concrete uses
$\lambda=10^{-2}$, temperature $2/3$, initial open probability 0.9,
stretch limits $(-0.1,1.1)$, and gate learning rate $10^{-2}$. STE uses
$\lambda=10^{-4}$, constant $\beta=2$, and boundary learning rate 0.05.

\subsection{Section~\ref{sec:experiments}: transferred structures}
For parallel feature, convolutional, and graph branches (GCNs in the graph
experiment \citep{kipf2017gcn}), the learned block is
\begin{equation}
  y(x)=h\!\left(\sum_{k=0}^{K-1}w_k g_k(t)f_k(x)\right),
  \label{eq:settings-parallel}
\end{equation}
where $h$ is the fixed task head.  In a Transformer layer
\citep{vaswani2017attention}, NGN instead gates
complete attention heads and FFN slices,
\begin{align}
  \operatorname{Attn}_{\ell}(H)
   & =\sum_k w_k g_k(t_\ell)
  \operatorname{softmax}\!\left(
  \frac{(HW^Q_{\ell k})(HW^K_{\ell k})^\top}{\sqrt{d_h}}\right)
  HW^V_{\ell k}W^O_{\ell k}, \\
  \operatorname{FFN}_{\ell}(h)
   & =\sum_k w_k g_k(t_\ell)
  W^{(2)}_{\ell k}\phi(W^{(1)}_{\ell k}h+b^{(1)}_{\ell k}).
  \label{eq:settings-transformer}
\end{align}
Here $d_h$ is the dimension of one attention head.
The Mamba analogue \citep{gu2023mamba} is
$\operatorname{Mamba}_{\ell}(H)=\sum_k w_k g_k(t_\ell)
  \operatorname{SSM}_{\ell k}(H)$, with one selective-state coordinate per term.

\paragraph{Learned depth.}
The synthetic input is $x=(x_1,x_2)\sim\mathcal N(0,I_2)$. Writing
$r=\sqrt{x_1^2+x_2^2}$ and $\theta=\operatorname{atan2}(x_2,x_1)$, the
three-step teacher used in Table~\ref{tab:data9} applies the radius-dependent
twist
\begin{equation}
  T(x)=r\bigl(\cos(\theta+3\gamma r),\,
  \sin(\theta+3\gamma r)\bigr),
  \qquad \gamma=0.5.
  \label{eq:twist-task}
\end{equation}
We draw 4,096 inputs, standardize each target coordinate over the generated
sample, and minimize mean squared error. The teacher expression is equivalent
to composing three rotations whose one-step angle is $\gamma r$.

The per-layer model assigns a separate boundary to each of $L=10$ layers:
\begin{equation}
  h_0=x, \qquad
  h_{\ell+1}=\sum_{k=0}^{K-1}w_{\ell k}g_k(t_\ell)
  f_{\ell k}(h_\ell), \quad \ell=0,\ldots,L-1.
  \label{eq:depth-per-layer}
\end{equation}
Here $f_{\ell k}(h)=\tanh(W_{\ell k}h+b_{\ell k})$; $f_{\ell0}$ is the
identity in the first $L-1$ layers, allowing a learned count of zero to remove
that transformation, while the final layer's $f_{L-1,0}$ is learned. The
recursive model instead shares one boundary across a chain:
\begin{equation}
  u_0=F_0(x), \qquad
  u_k=F_k(x+u_{k-1}), \qquad
  y=\sum_{k=0}^{K-1}w_k g_k(t)u_k,
  \label{eq:depth-recursive}
\end{equation}
where $F_k(v)=\tanh\!\left(V_k\tanh(U_kv+a_k)+b_k\right)$ is a width-32
residual MLP update. Thus the first construction learns a count at every layer,
whereas the second learns the length of one recursively composed chain.

\paragraph{Parameter-efficient adaptation and pruning.}
The large-model experiments adapt Llama-3-8B-Instruct
\citep{grattafiori2024llama3}. LoRA and adapters
\citep{hu2022lora,houlsby2019adapters} instantiate
$y=Wx+\sum_k w_k g_k(t)B_kA_kx$ and
$y=h+\sum_k w_k g_k(t)U_k\phi(D_kh)$, respectively.  Structured pruning
uses the same sum after partitioning attention heads or FFN neurons into
ordered blocks. For LoRA and adapters, $\delta=-0.5$, so hard deployment keeps
$\operatorname{round}(t)$ real blocks and permits an empty prefix at $t=0$.

\begin{table}[H]
  \centering
  \small
  \caption{Structural hyperparameters for Section~\ref{sec:experiments}.
    ``Per layer'' means that each listed layer has its own count boundary;
    delayed-linear $\alpha$ schedules begin at 50\% unless noted. A range
    $(a\text{--}b)$ means that the ramp starts at training fraction $a$ and
    reaches full strength at fraction $b$; a single value gives the start
    fraction and ramps through the remainder of training.}
  \label{tab:settings-experiments}
  \resizebox{\textwidth}{!}{%
    \begin{tabular}{lccccccc}
      \toprule
      task                  & $K$        & $\lambda_0$          & $\lambda$ sched. & $\alpha_{\max}$      & $\alpha$ sched.          & $\beta$      & $w_k$                 \\
      \midrule
      MNIST MLP             & 32         & $10^{-4}$            & power-decay      & $10^{-4}$            & delayed linear           & $4\!\to\!12$ & $0.5^k$               \\
      MNIST / CIFAR-10 CNN  & 8 / 16     & $10^{-4}$            & power-decay      & $10^{-3}$            & delayed linear           & $2\!\to\!8$  & $0.5^k$               \\
      Cora                  & 8          & $10^{-4}$            & power-decay      & $10^{-1}$            & delayed linear           & $4\!\to\!12$ & $0.5^k$               \\
      TinyStories attention & 8/layer    & $10^{-4}$            & power-decay      & $10^{-2}$            & delayed linear           & $2\!\to\!12$ & $0.5^k$               \\
      TinyStories FFN       & 16/layer   & $5\!\times\!10^{-4}$ & power-decay      & $5\!\times\!10^{-4}$ & delayed linear           & $4\!\to\!12$ & $0.5^k$               \\
      TinyStories joint     & 8/16/layer & $5\!\times\!10^{-4}$ & power-decay      & $5\!\times\!10^{-2}$ & delayed linear           & $2\!\to\!12$ & $0.5^k$               \\
      TinyStories Mamba     & 16/layer   & $10^{-4}$            & power-decay      & $10^{-2}$            & delayed linear           & $2\!\to\!12$ & $0.5^k$               \\
      twist depth           & 10/layer   & $10^{-3}$            & power-decay      & $10^{-3}$            & linear                   & $4\!\to\!12$ & Table~\ref{tab:data9} \\
      recursive twist       & 16         & $10^{-4}$            & power-decay      & $10^{-4}$            & linear                   & $4\!\to\!12$ & $1$                   \\
      Llama-3 LoRA          & 16/layer   & $10^{-3}$            & power-decay      & $10^{-2}$            & delayed linear (.5--.75) & $4\!\to\!12$ & $0.5^k$               \\
      Llama-3 adapter       & 16/layer   & $10^{-3}$            & power-decay      & $0.05$--$0.10$       & delayed linear (.75)     & $4\!\to\!12$ & $0.5^k$               \\
      Llama-3 pruning       & 16/layer   & $10^{-3}$            & constant         & $3\!\times\!10^{-2}$ & delayed linear (.75)     & $4\!\to\!12$ & $1$                   \\
      \bottomrule
    \end{tabular}}
\end{table}

All experiments other than large-model LoRA, adaptive-rank baselines, adapters,
and pruning ran on an NVIDIA RTX 4080 laptop GPU. TinyStories runs took roughly
one hour, the ten-layer depth study roughly 40 minutes, and the remaining small
experiments less than five minutes per seed. The large-model experiments ran on
an NVIDIA A100. Recorded large-model optimization times span 9--27 minutes and
end-to-end runs, including evaluation, span 10--98 minutes. The older Pythia
frontier files do not contain timers.
}

\newcommand{\DatasetLicensesAppendix}{%
  \section{Dataset licenses}
  \label{app:dataset-licenses}
  Table~\ref{tab:dataset-licenses} records the terms stated by the original
  distribution or the exact repository used in our experiments. MNIST, CIFAR-10,
  and Cora \citep{lecun1998mnist,krizhevsky2009cifar,yang2016planetoid} do not
  state an explicit dataset license on the cited official distribution pages; we
  therefore report this as unspecified rather than inferring a software license.
  The remaining external datasets are TinyStories, WikiSQL, Spider, GSM8K, SQuAD,
  and the Pile \citep{eldan2023tinystories,zhong2017wikisql,yu2018spider,
    cobbe2021gsm8k,rajpurkar2016squad,gao2020pile}. Synthetic 1D, twist, and
  manifold data are generated by our scripts and contain no third-party examples.
  \begin{table}[H]
    \centering
    \small
    \caption{Licenses and distribution sources for all datasets used in the
      reported experiments and forgetting evaluations.}
    \label{tab:dataset-licenses}
    \begin{tabular}{p{0.15\textwidth}p{0.27\textwidth}p{0.44\textwidth}}
      \toprule
      dataset        & use                       & stated license / source                                                                                                                                                \\
      \midrule
      MNIST          & image classification      & not specified on the \href{https://yann.lecun.com/exdb/mnist/}{original distribution page}                                                                             \\
      CIFAR-10       & image classification      & not specified on the \href{https://www.cs.toronto.edu/~kriz/cifar.html}{original distribution page}                                                                    \\
      Cora           & graph node classification & not specified by the \href{https://pytorch-geometric.readthedocs.io/en/latest/generated/torch_geometric.datasets.Planetoid.html}{Planetoid distribution documentation} \\
      TinyStories    & language modeling         & CDLA-Sharing-1.0 on the \href{https://huggingface.co/datasets/roneneldan/TinyStories}{dataset card}                                                                    \\
      WikiSQL        & text-to-SQL               & BSD-3-Clause in the \href{https://github.com/salesforce/WikiSQL}{official repository}                                                                                  \\
      Spider         & text-to-SQL               & CC BY-SA 4.0 on the \href{https://huggingface.co/datasets/xlangai/spider}{dataset card}                                                                                \\
      GSM8K          & mathematical reasoning    & MIT on the \href{https://huggingface.co/datasets/openai/gsm8k}{dataset card}                                                                                           \\
      SQuAD          & question answering        & CC BY-SA 4.0 on the \href{https://huggingface.co/datasets/rajpurkar/squad}{dataset card}                                                                               \\
      Pile-10k       & forgetting evaluation     & BigScience BLOOM RAIL 1.0 on the \href{https://huggingface.co/datasets/NeelNanda/pile-10k}{mirror card}; source components retain their own terms                      \\
      synthetic data & 1D, twist, manifold       & generated by the authors; no external dataset license                                                                                                                  \\
      \bottomrule
    \end{tabular}
  \end{table}
}

% Detailed adaptive-rank baseline allocations are kept out of the main table.
\newcommand{\AdaptiveRankBaselinesAppendix}{%
  \section{Adaptive-rank learned counts}
  \label{app:adaptive-rank-baselines}
  SoRA learns sparse gates over LoRA directions \citep{ding2023sora}, while
  AdaLoRA reallocates a prescribed rank budget \citep{zhang2023adalora}. A
  kept/target pair is shown when a run misses its requested budget. All WikiSQL
  LoRA runs use 10 epochs, while the WikiSQL adapter runs use 20 epochs. For NGN, the
  reported size is retained rank summed across layers. For SoRA, count is
  the number of directions whose independently learned gates survive proximal
  thresholding; AdaLoRA instead reports achieved versus prescribed target rank.
  The matched fixed NGN controls reproduce each layer's rounded learned
  allocation, including zero-capacity layers; the complete per-layer vectors are
  stored in the supplementary machine-readable results, while the tables report
  compact early/middle/late sums.
  \begin{table}[H]
    \centering
    \small
    \caption{Full NGN and matched-fixed diagnostics for Table~\ref{tab:data11}.
      Size is split across early/middle/late layer thirds; ``integer'' is the
      percentage of NGN boundaries within $0.01$ of an integer. Perplexity is
      measured on Pile relative to the base model.}
    \label{tab:data11-diagnostics}
    \resizebox{\textwidth}{!}{%
      \begin{tabular}{lcccccccc}
        \toprule
        task & base & size & e/m/l & integer & NGN soft/trunc. & fixed & NGN ppl/base & fixed ppl/base \\
        \midrule
        \dataelevenbody
        \bottomrule
      \end{tabular}}
  \end{table}

  \begin{table}[H]
    \centering
    \small
    \caption{Detailed NGN, SoRA, and AdaLoRA results corresponding to the compact score
      columns in Table~\ref{tab:data11}. Size is total retained rank, e/m/l splits
      it across early/middle/late layer thirds, and forgetting is Pile perplexity
      relative to the base model. GSM8K scores use the same flexible numerical
      extraction exact match as Table~\ref{tab:data11}.}
    \label{tab:data11-baselines}
    \begin{tabular}{llcccc}
      \toprule
      task & method & learned count & e/m/l & score & ppl$/$base \\
      \midrule
      \dataelevenbaselinesbody
      \bottomrule
    \end{tabular}
  \end{table}
}

% Full uncertainty tables are kept out of the main text and invoked after
% \appendix by the paper and standalone preview.
\newcommand{\UncertaintyTablesAppendix}{%
  \section{Additional data with standard deviations}
  \label{app:full-uncertainty}

  \begin{table}[H]
    \centering
    \caption{Full schedule-ablation results corresponding to
      Table~\ref{tab:data4}; mean$\,\pm\,$population standard deviation over ten
      seeds. The baseline MSE statistics are dominated by one high-error seed
      ($7.82\times10^{-3}$); the median baseline MSE is $1.31\times10^{-5}$.}
    \label{tab:data4-full}
    \begin{tabular}{lccc}
      \toprule
      setting & $t$ & soft MSE & truncated MSE \\
      \midrule
      \datafourfullbody
      \bottomrule
    \end{tabular}
  \end{table}

  \begin{table}[H]
    \centering
    \small
    \caption{Full results for the non-language-model rows of
      Table~\ref{tab:data8}; mean$\,\pm\,$population standard deviation over ten
      seeds. Fixed models are retrained independently at each seed's rounded
      learned boundary. The final column reports the aggregate label followed by
      the prune-from-high boundary and counts of yes/partial/no seed outcomes
      (Y/P/N), using the definitions in Table~\ref{tab:data8}. Values
      are displayed to two decimal places; the MNIST MLP entry
      $8.00\,\pm\,0.00$ has unrounded mean $7.99955$ and standard deviation
      $0.00049$, reflecting consistent convergence to the integer boundary rather
      than identical runs.}
    \label{tab:data8-full}
    \setlength{\tabcolsep}{3pt}
    \resizebox{\textwidth}{!}{%
      \begin{tabular}{llccccl}
        \toprule
        task & $f_k$ & learned $t$ & soft acc. & trunc. acc. & fixed acc. & prunes back? \\
        \midrule
        \dataeightfullbody
        \bottomrule
      \end{tabular}}
  \end{table}
}

\newcommand{\ManifoldSetupAppendix}{%
\section{Nonlinear manifold-dimension setup}
\label{app:manifold-setup}

For each $D\in\{1,2,4,6,8\}$, we draw $z$ uniformly from $[-1,1]^D$, append
$16-D$ zeros, apply a fixed random orthogonal rotation $R$, and map the result
coordinatewise through an invertible nonlinearity,
\begin{equation}
  \phi(z)=\tanh\!\left(0.8R[z;0]\right).
\end{equation}
This gives a smooth injective nonlinear image of a $D$-dimensional box in the
16-dimensional ambient space while keeping its intrinsic dimension known.

The autoencoder uses four width-64 GELU layers in the encoder and four in the
decoder. Between them is a variational sum of ten zero-initialized rank-one
terms controlled by one soft NGN boundary, with no skip around the bottleneck.
Writing $\mu_k(x)=A_kE(x)$ and letting $\ell_k$ be a learned log variance, the
scalar supplied to rank-one direction $B_k$ during training is sampled from
\begin{equation}
  q_k(z_k\mid x)=\mathcal N\!\left(
  g_k(t)\mu_k(x),\;
  g_k(t)^2\exp(\ell_k)+1-g_k(t)^2
  \right).
\end{equation}
A closed gate therefore approaches input-independent unit Gaussian noise,
preventing a decoder from recovering hidden information merely by amplifying
a small gate. We add $10^{-4}\sum_k
  D_{\mathrm{KL}}(q_k(z_k\mid x)\Vert\mathcal N(0,1))$ to the objective and use
posterior means for evaluation.

Each seed takes 50,000 Adam steps with fresh batches of 1,024 points and is
evaluated on 2,048 held-out points. The body learning rate is $10^{-3}$ with cosine decay,
the boundary learning rate is $5\times10^{-2}$, gradient clipping is 1,
$w_k=0.5^k$, $t_0=0$, and $\delta=0.5$. We use
$\lambda_0=2\times10^{-3}$ and $\alpha_{\max}=10^{-3}$:
$\lambda$ decreases linearly to zero,
$\alpha$ begins its linear ramp halfway through training, and $\beta$ increases
linearly from 4 to 12. Reconstruction loss averages over samples and sums the 16
coordinate errors. No STE or hard gate is used during training.

\begin{table}[H]
  \centering
  \small
  \caption{Full manifold-dimension results corresponding to
    Table~\ref{tab:data15}; mean$\,\pm\,$population standard deviation over ten
    seeds. The NMSE distributions are right-skewed: three $D=2$ seeds and one
    $D=4$ seed have visible soft--truncated gaps, while one $D=8$ seed selects
    seven rather than eight terms and reaches truncated NMSE
    $9.81\times10^{-3}$.}
  \label{tab:data15-full}
  \begin{tabular}{rcccc}
    \toprule
    true $D$ & learned $D$ & exact rate & soft NMSE & trunc. NMSE \\
    \midrule
    \datafifteenfullbody
    \bottomrule
  \end{tabular}
\end{table}
}

% Kept here with the experiment data; invoked after \appendix by the paper
% and by the standalone preview. This avoids placing later experiments in appendix.
\newcommand{\DecayAblationAppendix}{%
  \section{Learned-decay and normalization ablations: AWNN components}
  \label{sec:decay}
  \label{app:decay-ablation}
  AWNN combines three relevant ingredients: a learned geometric importance decay,
  normalization of the weighted functions, and a quantile rule that converts the
  importance distribution into a width. Table~\ref{tab:decay-ablation} isolates
  the first two ingredients on the same MLP function bank, while the complete AWNN
  implementation in Figure~\ref{fig:data7} additionally applies the quantile width
  during training. These are component ablations, not a claim to reproduce every
  architectural detail of the published model.

  In NGN notation, AWNN maps naturally to replacing the fixed
  $w_k=0.5^k$ by $w_k=r^k$ with learned $r=\sigmoid(\rho)$, normalizing each
  function as $f_k/\lVert f_k\rVert$, removing the learned boundary $t$, and
  deploying the smallest prefix containing 90\% of the finite geometric mass.
  The rows below progressively make $r$ learnable, retain or remove $t$, and turn
  function normalization on or off. This puts the AWNN ingredients inside the
  same NGN function bank so their effects can be separated.
  \begin{table}[H]
    \centering
    \caption{Ablations of the learned-decay and normalization components used by
      AWNN, evaluated on the common MLP function bank. Alternative rows use ten
      seeds; the standard row is a single historical run. Its fixed-MSE cell is
      the hard MSE, not an independent retraining result.}
    \label{tab:decay-ablation}
    \setlength{\tabcolsep}{4pt}
    \resizebox{\textwidth}{!}{%
      \begin{tabular}{lccccc}
        \toprule
        setting & $r$ & $t$ & soft MSE & hard MSE & fixed MSE \\
        \midrule
        \decayablationbody
        \bottomrule
      \end{tabular}}
  \end{table}
  The decomposition explains the failure mode observed here. Learning $r$ alone
  can obtain a strong soft fit by spreading useful signal through the full tail,
  but the post-hoc quantile cutoff removes part of that signal. RMS normalization
  then divides out much of the amplitude ordering induced by a positive decay
  coefficient, weakening the mechanism that is supposed to make later functions
  dispensable. The complete AWNN baseline selects a stable width, but retains the
  higher deployed error visible in Figure~\ref{fig:data7}. This is a failure on the
  controlled task, not a general claim that AWNN cannot work on its native
  architectures. NGN avoids this particular mismatch by optimizing the boundary
  used for hard deployment and explicitly snapping it to an integer.
}

%% file: data_generated.tex
% AUTO-GENERATED by build_paper_data.py -- do not edit by hand.

\newcommand{\stdT}{7.00}
\newcommand{\stdSoftMSE}{6.27e-06}
\newcommand{\stdHardMSE}{6.28e-06}
\newcommand{\datafourteenintrate}{34\%}
\newcommand{\datafourteenmixintrate}{100\%}

\newcommand{\datafourbody}{%
\multicolumn{5}{l}{\emph{Schedule robustness}} \\
baseline & 6.00 & 100\% & 1.31e-05 & 1.31e-05 \\
$\lambda$ const & 5.45 & 0\% & 1.84e-05 & 1.84e-05 \\
$\alpha=10^{-4}$ const & 7.00 & 100\% & 1.41e-05 & 1.41e-05 \\
\addlinespace
\multicolumn{5}{l}{\emph{Component ablations}} \\
$\alpha=0$ & 4.49 & 0\% & 5.43e-05 & 4.83e-04 \\
$\beta=4$ const & 6.00 & 100\% & 1.50e-05 & 6.01e-04 \\
$\beta=15$ const & 2.00 & 90\% & 7.54e-04 & 7.54e-04 \\
}

\newcommand{\datafourfullbody}{%
\multicolumn{4}{l}{\emph{Schedule robustness}} \\
baseline & 6.10\,$\pm$\,1.45 & 9.32e-04\,$\pm$\,2.3e-03 & 9.32e-04\,$\pm$\,2.3e-03 \\
$\lambda$ const & 5.75\,$\pm$\,1.83 & 3.12e-04\,$\pm$\,4.9e-04 & 3.12e-04\,$\pm$\,4.9e-04 \\
$\alpha=10^{-4}$ const & 7.30\,$\pm$\,0.78 & 2.24e-04\,$\pm$\,3.2e-04 & 2.24e-04\,$\pm$\,3.2e-04 \\
\addlinespace
\multicolumn{4}{l}{\emph{Component ablations}} \\
$\alpha=0$ & 4.70\,$\pm$\,1.49 & 3.25e-04\,$\pm$\,4.9e-04 & 1.45e-03\,$\pm$\,2.6e-03 \\
$\beta=4$ const & 6.30\,$\pm$\,1.42 & 3.09e-04\,$\pm$\,5.2e-04 & 6.46e-04\,$\pm$\,5.7e-04 \\
$\beta=15$ const & 2.05\,$\pm$\,0.57 & 1.73e-03\,$\pm$\,3.0e-03 & 1.73e-03\,$\pm$\,3.0e-03 \\
}

\newcommand{\datasevenbody}{%
AWNN (block adaptation) & 6.54 & 5.43e-04 & 5.43e-04 & 2.72e-04 \\
STE & 5.46 & 8.01e-04 & 8.01e-04 & 2.91e-05 \\
$L_0$ (hard concrete) & 8.38 & 5.08e-04 & 5.08e-04 & 9.75e-05 \\
NGN (baseline) & 6.28 & 2.51e-04 & 2.51e-04 & -- \\
}

\newcommand{\datasevenfullbody}{%
AWNN (block adaptation) & 6.54\,$\pm$\,0.61 & 5.43e-04\,$\pm$\,1.5e-04 & 5.43e-04\,$\pm$\,1.5e-04 & 2.72e-04\,$\pm$\,1.5e-04 \\
STE & 5.46\,$\pm$\,1.63 & 8.01e-04\,$\pm$\,1.8e-03 & 8.01e-04\,$\pm$\,1.8e-03 & 2.91e-05\,$\pm$\,3.2e-05 \\
$L_0$ (hard concrete) & 8.38\,$\pm$\,1.15 & 5.08e-04\,$\pm$\,7.9e-04 & 5.08e-04\,$\pm$\,7.9e-04 & 9.75e-05\,$\pm$\,1.3e-04 \\
NGN (baseline) & 6.28\,$\pm$\,1.46 & 2.51e-04\,$\pm$\,1.1e-03 & 2.51e-04\,$\pm$\,1.1e-03 & -- \\
}

\newcommand{\decayablationbody}{%
standard ($r=0.5$, learn $t$) & 0.50 & 7.00 & 6.27e-06 & 6.28e-06 & 6.28e-06 \\
learn $r$ and $t$ & 0.81\,$\pm$\,0.04 & 5.77\,$\pm$\,1.42 & 4.84e-05\,$\pm$\,5.7e-05 & 3.01e-02\,$\pm$\,6.3e-02 & 8.59e-05\,$\pm$\,2.1e-04 \\
learn $t$, norm., fixed $r=0.5$ & 0.50 & 3.42\,$\pm$\,1.32 & 6.36e-02\,$\pm$\,4.7e-02 & 9.03e-02\,$\pm$\,1.1e-01 & 2.30e-05\,$\pm$\,2.6e-05 \\
learn $r$ only & 0.81\,$\pm$\,0.02 & -- & 1.26e-06\,$\pm$\,9.8e-07 & 8.23e+00\,$\pm$\,2.3e+00 & 4.12e-05\,$\pm$\,8.5e-05 \\
learn $r$, normalized & 0.89\,$\pm$\,0.04 & -- & 2.75e-02\,$\pm$\,1.4e-02 & 2.75e-01\,$\pm$\,1.5e-01 & 4.05e-05\,$\pm$\,7.8e-05 \\
}

\newcommand{\dataeightbody}{%
MNIST & tanh MLP features & \texttt{[8.00]} & 100\% & 96.93 & 96.93 & 96.81 & yes \\
MNIST & conv branches & \texttt{[4.00]} & 100\% & 99.18 & 99.18 & 99.13 & partial \\
CIFAR-10 & conv branches & \texttt{[7.02]} & 50\% & 71.77 & 71.77 & 72.03 & partial \\
Cora & 2-hop GCN branches & \texttt{[1.00]} & 100\% & 79.05 & 79.05 & 79.45 & partial \\
TinyStories & attention heads ($\times 2$ layers) & \texttt{[2.11, 3.03]} & 0\% & 1.3055 & 1.3234 & 1.3011 & no \\
TinyStories & FFN blocks ($\times 2$ layers) & \texttt{[1.14, 8.00]} & 50\% & 1.3454 & 1.3775 & 1.3402 & partial \\
TinyStories & heads + FFN blocks ($\times 2$ layers) & \texttt{[1.01, 3.02, 1.01, 10.00]} & 50\% & 1.3553 & 1.3902 & 1.3270 & no \\
TinyStories & Mamba state dim. $N$ ($\times 2$ layers) & \texttt{[1.00, 2.00]} & 100\% & 1.3551 & 1.3597 & 1.3445 & partial \\
}

\newcommand{\dataeightfullbody}{%
MNIST & tanh MLP features & \texttt{[8.00\,$\pm$\,0.00]} & 96.94\,$\pm$\,0.07 & 96.94\,$\pm$\,0.07 & 96.81\,$\pm$\,0.09 & yes (\texttt{[8.20\,$\pm$\,0.40]}; Y/P/N=10/0/0) \\
MNIST & conv branches & \texttt{[4.10\,$\pm$\,0.30]} & 99.17\,$\pm$\,0.03 & 99.17\,$\pm$\,0.03 & 99.14\,$\pm$\,0.04 & partial (\texttt{[5.60\,$\pm$\,0.49]}; Y/P/N=5/4/1) \\
CIFAR-10 & conv branches & \texttt{[7.41\,$\pm$\,0.49]} & 71.74\,$\pm$\,0.26 & 71.74\,$\pm$\,0.26 & 72.08\,$\pm$\,0.29 & partial (\texttt{[10.00\,$\pm$\,0.00]}; Y/P/N=0/10/0) \\
Cora & 2-hop GCN branches & \texttt{[1.20\,$\pm$\,0.40]} & 78.68\,$\pm$\,1.66 & 78.68\,$\pm$\,1.66 & 79.43\,$\pm$\,1.18 & partial (\texttt{[2.60\,$\pm$\,0.49]}; Y/P/N=5/5/0) \\
}

\newcommand{\dataninebody}{%
per-layer: const-$\lambda$, const-$r$ & \texttt{[3,4,4,3,3,1,3,2,2,6]} & 10 & 2.28e-02 & 2.28e-02 & 1.72e-02 \\
per-layer: const-$\lambda$, inc-$r$ & \texttt{[4,2,3,3,1,3,2,3,4,9]} & 10 & 7.37e-04 & 7.74e-04 & 6.98e-04 \\
per-layer: dec-$\lambda$, const-$r$ & \texttt{[0,0,0,4,5,5,5,8,9,9]} & 7 & 1.57e-02 & 1.58e-02 & 3.28e-02 \\
per-layer: dec-$\lambda$, inc-$r$ & \texttt{[0,0,0,0,2,3,7,9,9,9]} & 6 & 2.51e-04 & 3.30e-04 & 3.78e-03 \\
}

\newcommand{\datatenbody}{%
recursive & 7.02 & 8 & 2.49e-05 & 3.56e-05 & 1.54e-05 \\
}

\newcommand{\dataelevenlorabody}{%
WikiSQL-easy & 3.5 & 19 (2/9/8) & 78.2 / 78.6 & 77.7 & 75.2 & 45.1 \\
WikiSQL-hard & 2.0 & 21 (4/11/6) & 70.8 / 70.4 & 68.7 & 67.4 & 45.4 \\
Spider & 1.1 & 17 (5/7/5) & 12.6 / 12.4 & 12.3 & 13.9 & 10.3 \\
GSM8K & 45.2 & 18 (7/8/3) & 53.6 / 55.2 & 50.8 & 35.6 & 56.8 \\
SQuAD & 65.2 & 13 (2/7/4) & 67.2 / 67.3 & 67.1 & 67.1 & 66.9 \\
}

\newcommand{\dataelevenadapterbody}{%
WikiSQL-easy & 3.5 & 4 (2/1/1) & 83.2 / 80.5 & 82.4 \\
WikiSQL-hard & 2.0 & 6 (3/1/2) & 73.1 / 73.2 & 72.9 \\
Spider & 1.1 & 5 (1/2/2) & 13.5 / 11.0 & 14.6 \\
GSM8K & 45.2 & 6 (2/3/1) & 55.4 / 52.8 & 58.6 \\
SQuAD & 65.2 & 1 (0/1/0) & 69.9 / 66.8 & 71.9 \\
}

\newcommand{\dataelevenbody}{%
\multicolumn{9}{l}{\emph{LoRA rank}} \\
WikiSQL-easy & 3.5 & 19 & 2/9/8 & 20\% & 78.2 / 78.6 & 77.7 & 1.016 & 0.996 \\
WikiSQL-hard & 2.0 & 21 & 4/11/6 & 28\% & 70.8 / 70.4 & 68.7 & 1.022 & 0.999 \\
Spider & 1.1 & 17 & 5/7/5 & 58\% & 12.6 / 12.4 & 12.3 & 0.997 & 0.984 \\
GSM8K & 45.2 & 18 & 7/8/3 & 62\% & 53.6 / 55.2 & 50.8 & 0.968 & 0.973 \\
SQuAD & 65.2 & 13 & 2/7/4 & 34\% & 67.2 / 67.3 & 67.1 & 0.993 & 0.980 \\
\midrule
\multicolumn{9}{l}{\emph{Adapter width}} \\
WikiSQL-easy & 3.5 & 4 & 2/1/1 & 44\% & 83.2 / 80.5 & 82.4 & 1.018 & 1.006 \\
WikiSQL-hard & 2.0 & 6 & 3/1/2 & 47\% & 73.1 / 73.2 & 72.9 & 1.031 & 1.002 \\
Spider & 1.1 & 5 & 1/2/2 & 12\% & 13.5 / 11.0 & 14.6 & 1.000 & 0.995 \\
GSM8K & 45.2 & 6 & 2/3/1 & 47\% & 55.4 / 52.8 & 58.6 & 0.968 & 0.969 \\
SQuAD & 65.2 & 1 & 0/1/0 & 16\% & 69.9 / 66.8 & 71.9 & 0.995 & 0.997 \\
}

\newcommand{\dataelevenbaselinesbody}{%
WikiSQL-easy & NGN & 19 & 2/9/8 & 78.6 & 1.016 \\
WikiSQL-easy & SoRA & 19 & 7/12/0 & 75.2 & 1.009 \\
WikiSQL-easy & AdaLoRA & 19 & 3/13/3 & 45.1 & 1.087 \\
WikiSQL-hard & NGN & 21 & 4/11/6 & 70.4 & 1.022 \\
WikiSQL-hard & SoRA & 21 & 9/12/0 & 67.4 & 1.000 \\
WikiSQL-hard & AdaLoRA & 21 & 5/14/2 & 45.4 & 1.063 \\
Spider & NGN & 17 & 5/7/5 & 12.4 & 0.997 \\
Spider & SoRA & 13/17 & 7/6/0 & 13.9 & 0.999 \\
Spider & AdaLoRA & 17 & 6/9/2 & 10.3 & 1.023 \\
GSM8K & NGN & 18 & 7/8/3 & 55.2 & 0.968 \\
GSM8K & SoRA & 11/18 & 4/7/0 & 35.6 & 0.969 \\
GSM8K & AdaLoRA & 18 & 11/7/0 & 56.8 & 0.987 \\
SQuAD & NGN & 13 & 2/7/4 & 67.3 & 0.993 \\
SQuAD & SoRA & 13 & 4/4/5 & 67.1 & 0.994 \\
SQuAD & AdaLoRA & 13 & 4/5/4 & 66.9 & 1.006 \\
}

\newcommand{\datathirteenbody}{%
attention heads & frozen & $10^{-2}$ & 78.0\% & 76.3/79.7/77.9 & 41.5 & 9.61 \\
attention heads & frozen & $10^{-3}$ & 91.2\% & 91.4/91.9/90.4 & 50.5 & 1.80 \\
attention heads & frozen & $10^{-4}$ & 94.9\% & 94.0/94.3/96.4 & 57.0 & 1.50 \\
attention heads & updated & $10^{-2}$ & 72.9\% & 69.8/76.0/72.9 & 37.0 & 561.01 \\
attention heads & updated & $10^{-3}$ & 84.7\% & 85.2/86.7/82.3 & 54.5 & 219.20 \\
attention heads & updated & $10^{-4}$ & 89.3\% & 89.1/92.2/86.7 & 63.0 & 220.37 \\
FFN neurons & frozen & $10^{-2}$ & 52.4\% & 62.5/53.1/41.7 & 40.5 & 184.03 \\
FFN neurons & frozen & $10^{-3}$ & 84.4\% & 88.5/83.3/81.2 & 56.5 & 3.30 \\
FFN neurons & frozen & $10^{-4}$ & 92.7\% & 93.2/91.7/93.2 & 64.0 & 1.76 \\
FFN neurons & updated & $10^{-2}$ & 42.9\% & 50.0/40.6/38.0 & 53.5 & 789.92 \\
FFN neurons & updated & $10^{-3}$ & 66.5\% & 74.5/67.2/57.8 & 56.5 & 75.07 \\
FFN neurons & updated & $10^{-4}$ & 83.7\% & 84.4/82.3/84.4 & 61.5 & 67.72 \\
both & frozen & $10^{-2}$ & 27.5\% & 33.3/26.8/22.4 & 33.0 & 665.22 \\
both & frozen & $10^{-3}$ & 76.1\% & 78.6/76.6/73.2 & 59.5 & 5.88 \\
both & frozen & $10^{-4}$ & 87.6\% & 88.0/85.9/88.8 & 64.5 & 2.83 \\
both & updated & $10^{-2}$ & 13.5\% & 13.5/14.8/12.0 & 27.5 & 887.92 \\
both & updated & $10^{-3}$ & 57.5\% & 62.8/59.9/49.7 & 51.5 & 249.87 \\
both & updated & $10^{-4}$ & 73.8\% & 75.8/75.8/69.8 & 61.0 & 100.01 \\
}

\newcommand{\datafourteenbody}{%
NGN prune & 14.1\% & 19.9/10.8/11.2 & 62.5 & 639 & 34.2\% & 36.4/35.8/30.0 & 70.0 & 42 \\
random drop & 14.1\% & 19.9/10.8/11.2 & 17.0 & 7.0e4 & 34.2\% & 36.4/35.8/30.0 & 17.5 & 1.6e3 \\
Wanda (per-layer) & 14.1\% & 19.9/10.8/11.2 & 69.5 & 356 & 34.2\% & 36.4/35.8/30.0 & 73.0 & 45 \\
}

\newcommand{\datafifteenbody}{%
1 & 1.00 & 100\% & 9.94e-07 & 9.94e-07 \\
2 & 2.00 & 100\% & 2.39e-06 & 2.39e-06 \\
4 & 4.00 & 100\% & 1.14e-05 & 1.14e-05 \\
6 & 6.00 & 100\% & 5.55e-05 & 5.55e-05 \\
8 & 8.00 & 90\% & 1.57e-04 & 1.57e-04 \\
}

\newcommand{\datafifteenfullbody}{%
1 & 1.00\,$\pm$\,0.00 & 100\% & 9.79e-07\,$\pm$\,5.7e-08 & 9.79e-07\,$\pm$\,5.7e-08 \\
2 & 2.00\,$\pm$\,0.00 & 100\% & 1.39e-05\,$\pm$\,1.5e-05 & 1.36e-03\,$\pm$\,2.5e-03 \\
4 & 4.00\,$\pm$\,0.00 & 100\% & 2.86e-05\,$\pm$\,3.8e-05 & 1.18e-04\,$\pm$\,3.0e-04 \\
6 & 6.00\,$\pm$\,0.00 & 100\% & 6.79e-05\,$\pm$\,4.3e-05 & 6.79e-05\,$\pm$\,4.3e-05 \\
8 & 7.90\,$\pm$\,0.30 & 90\% & 1.25e-03\,$\pm$\,2.9e-03 & 1.25e-03\,$\pm$\,2.9e-03 \\
}